%% file: formatting-instructions-latex-2020.tex
\documentclass{article}
\usepackage{ijcai20}

\usepackage{times}
\usepackage{soul}
\usepackage{url}
\usepackage[hidelinks]{hyperref}
\usepackage[utf8]{inputenc}
\usepackage[small]{caption}
\usepackage{graphicx}
\usepackage{amsmath}
\usepackage{amsthm}
\usepackage{booktabs}
\usepackage{algorithm}
\usepackage{algorithmic}
\usepackage{multirow}
\usepackage{amssymb,amsmath}

\newcommand{\modelname}{\textsc{SynonymNet}}

\title{Entity Synonym Discovery via Multipiece Bilateral Context Matching}

\author{
Chenwei Zhang$^1$\footnote{Work done while at the University of Illinois at Chicago}\and
Yaliang Li$^2$\and
Nan Du$^{3}$\and
Wei Fan$^{3}$\And
Philip S. Yu$^{4}$\\
\affiliations
$^1$Amazon, Seattle, WA 98109 USA\\
$^2$Alibaba Group, Bellevue, WA 98004 USA\\
$^3$Tencent Medical AI Lab, Palo Alto, CA 94306 USA\\
$^4$University of Illinois at Chicago, Chicago, IL 60607 USA
\emails
cwzhang@amazon.com,
yaliang.li@alibaba-inc.com,
\{ndu,davidwfan\}@tencent.com,
psyu@uic.edu
}

\begin{document}

\maketitle

\begin{abstract}
Being able to automatically discover synonymous entities in an open-world setting benefits various tasks such as entity disambiguation or knowledge graph canonicalization. Existing works either only utilize entity features, or rely on structured annotations from a single piece of context where the entity is mentioned. To leverage diverse contexts where entities are mentioned, in this paper, we generalize the distributional hypothesis to a multi-context setting and propose a synonym discovery framework that detects entity synonyms from free-text corpora with considerations on effectiveness and robustness.
As one of the key components in synonym discovery, we introduce a neural network model {\modelname} to determine whether or not two given entities are synonym with each other. Instead of using entities features, {\modelname} makes use of multiple pieces of contexts in which the entity is mentioned, and compares the context-level similarity via a bilateral matching schema. Experimental results demonstrate that the proposed model is able to detect synonym sets that are not observed during training on both generic and domain-specific datasets: Wiki+Freebase, PubMed+UMLS, and MedBook+MKG, with up to 4.16\% improvement in terms of Area Under the Curve and 3.19\% in terms of Mean Average Precision compared to the best baseline method. Code and data are available\footnote{\url{https://github.com/czhang99/SynonymNet}}.
\end{abstract}
\input{subfiles/1_intro.tex}
\input{subfiles/2_model.tex}
\input{subfiles/3_experiments.tex}

\input{subfiles/4_related.tex}
\input{subfiles/5_conclusion.tex}
\section*{Acknowledgments}
We thank the reviewers for their valuable comments. This work is supported in part by NSF under grants III-1526499, III-1763325, III-1909323, and CNS-1930941. 

\newpage
\bibliographystyle{named}
\bibliography{conference}
\end{document}

%% file: subfiles/1_intro.tex
\section{Introduction}
Discovering synonymous entities from a massive corpus is an indispensable task in automated knowledge discovery. For each entity, its synonyms refer to the entities that can be used interchangeably under certain contexts. For example, \texttt{clogged nose} and \texttt{nasal congestion} are synonyms relative to the context in which they are mentioned.
Given two entities, the synonym discovery task determines how likely these two entities are synonym with each other. 
The main goal of synonym discovery is to learn a metric that distinguishes synonym entities from non-synonym ones.

The synonym discovery task is challenging to deal with for the following reasons. First of all, entities are expressed with variations. For example, \texttt{U.S.A}/\texttt{United States of America}/\texttt{United States}/\texttt{U.S.} refer to the same idea but are expressed quite differently.
Recent works on synonym discovery focus on learning the similarity from entities and their character-level features \cite{neculoiu2016learning,mueller2016siamese}. These methods work well for synonyms that share a lot of character-level features like \texttt{airplane}/\texttt{aeroplane} or an entity and its abbreviation like \texttt{Acquired Immune Deficiency Syndrome}/\texttt{AIDS}. However, a large number of synonym entities in the real world do not share a lot of character-level features, such as \texttt{JD}/\texttt{law degree}, or \texttt{clogged nose} expressed on social media vs. \texttt{nasal congestion} mentioned in medical books. With only character-level features being used, these models hardly obtain the ability to discriminate entities that share similar semantics but are not alike verbatim. Secondly, the nature of synonym discovery tasks in real-world scenarios makes it common yet more difficult under an open-world setting: new entities and synonyms emerge and need to be discovered from the text corpora.

Context information helps indicate entity synonymity. The distributional semantics theory \cite{harris1954distributional,firth1957synopsis} hypothesizes that the meaning of an entity can be reflected by its neighboring words in the text.
Current works achieve decent performance on entity similarity learning, but still suffer from the following issues:
1) \noindent\textbf{Semantic Structure}. Context, as a snippet of natural language sentence, is semantically structured. Some existing models encode the semantic structures in contexts implicitly during the entity representation learning process \cite{mikolov2013distributed,pennington2014glove,peters2018deep}.
The entity representations embody meaningful semantics: entities with similar contexts are likely to live in proximity in the embedding space.
Some other works explicitly incorporate structured annotations to model contexts. 
Dependency parsing tree \cite{qu2017automatic}, user click information \cite{wei2009context}, or signed heterogeneous graphs \cite{ren2015synonym} are introduced as the structured information to help discover synonyms. However, structured annotations are time-consuming to obtain and may not even exist in an open-world setting.
2) \noindent\textbf{Diverse Contexts}.
A single entity can be mentioned in different contexts, let alone the case for multiple synonymous entities. Previous works on context-based synonym discovery either focus on entity information only \cite{neculoiu2016learning,mueller2016siamese}, or a single piece of context for each entity \cite{liao2017deep,qu2017automatic} for context matching. Notably, in specific domains such as medical, individuals(patients/doctors) may provide different context information when mentioning the same entity. Thus. using a single piece of context may suffer from noises.
Incorporating multiple pieces of contexts explicitly for entity matching has the potential to improve both accuracy and robustness, which is less studied in existing works. Moreover, it is not practical to assume that multiple pieces of contexts are equally informative to represent the meaning of an entity: a context may contribute differently when being matched to different entities. Thus it is imperative to focus on multiple pieces of contexts with a dynamic matching schema.

In light of these challenges, we propose a framework to discover synonym entities from a massive corpus without additional structured annotations.
A neural network model {\modelname} is proposed to detect entity synonyms based on two given entities via a bilateral matching among multiple pieces of contexts in which each entity appears. A leaky unit is designed to explicitly alleviate the noises from uninformative context during the matching process. We generate synonym entities that are completely unseen during training in the experiments.

The contribution of this work is summarized as follows:
\begin{itemize}
\item We propose {\modelname}, a context-aware bilateral matching model to detect entity synonyms. {\modelname} generalizes the distributional hypothesis to multiple pieces of contexts.
\item We introduce a synonym discovery framework that adopts {\modelname} to obtain synonym entities from a free-text corpus without additional annotation.
\item Experiments are conducted with an open-world setting on generic and domain-specific datasets in English and Chinese, which demonstrate the effectiveness of the proposed model for synonym discovery.
\end{itemize}

%% file: subfiles/2_model.tex
\section{\modelname}
\begin{figure}[bth!]
\centering
\includegraphics[width=\linewidth]{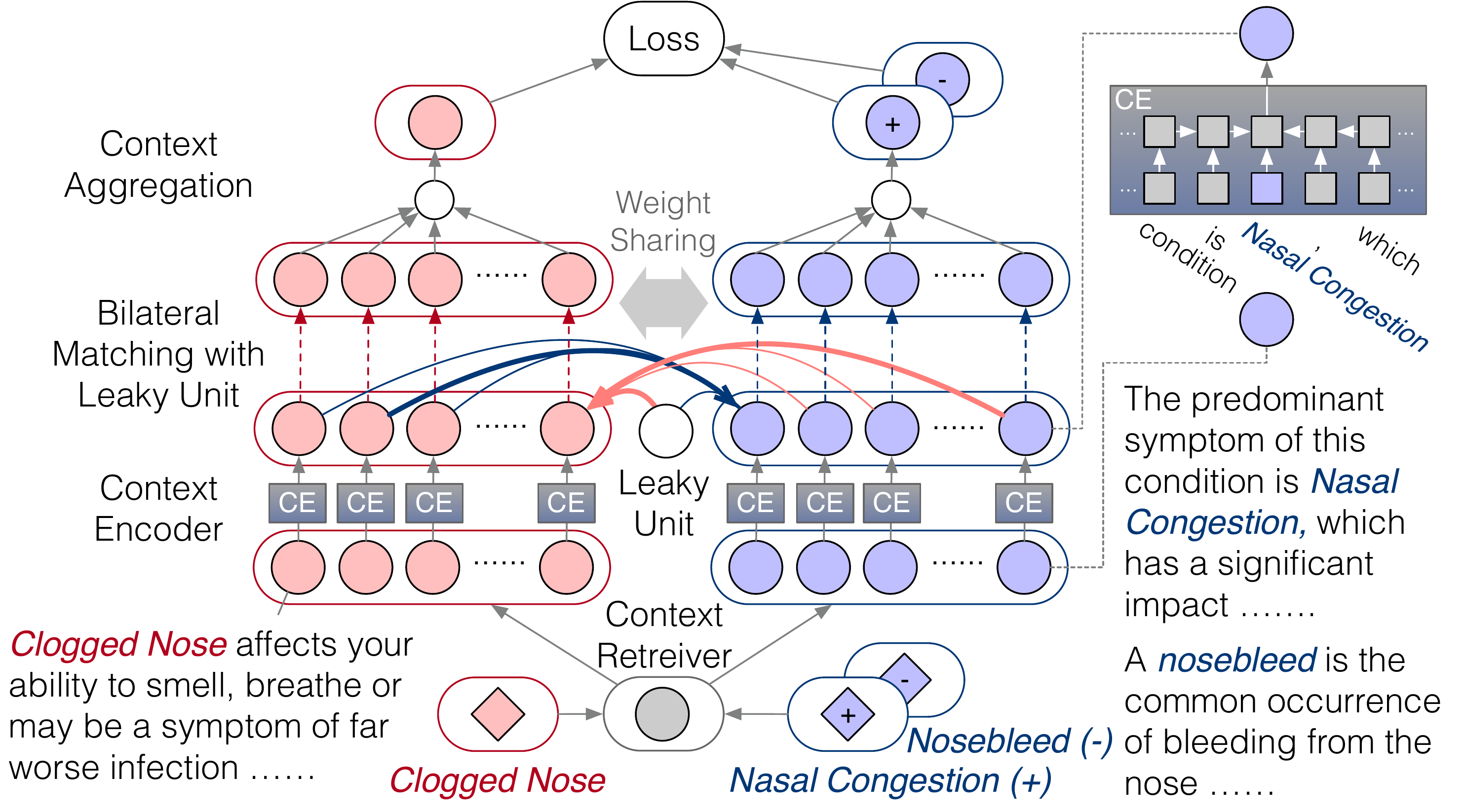}
\caption{Overview of the proposed model {\modelname}. The diamonds are entities. Each circle is associated with a piece of context in which an entity appears. {\modelname} learns to minimize the loss calculated using multiple pieces of contexts via bilateral matching with leaky units.}\label{fig::model_overview}
\end{figure}

We introduce {\modelname}, our proposed model that detects whether or not two entities are synonyms to each other based on a bilateral matching between multiple pieces of contexts in which entities appear.
Figure \ref{fig::model_overview} gives an overview of the proposed model.

\subsection{Context Retriever}
For each entity $e$, the context retriever randomly fetches $P$ pieces of contexts from the corpus $D$ in which the entity appears. We denote the retrieved contexts for $e$ as a set 
$C=\{c_1,...,c_P\}$, where $P$ is the number of context pieces.
Each piece of context $c_p \in C$ contains a sequence of words
$c_p = (w^{(1)}_{p}, ..., w^{(T)}_{p}),$
where $T$ is the length of the context, which varies from one instance to another. ${w}^{(t)}_{p}$ is the $t$-th word in the $p$-th context retrieved for an entity $e$.
\subsection{Context Encoder}
For the $p$-th context $c_p$, an encoder tries to learn a continuous vector that represents the context.
For example, a recurrent neural network (RNN) such as a bidirectional LSTM (Bi-LSTM) \cite{hochreiter1997long} can be applied to sequentially encode the context into hidden states:
\begin{equation}
\begin{aligned}
{\mathbf{\overset{\lower0.5em\hbox{$\smash{\scriptscriptstyle\rightarrow}$}}{h^{(t)}_p} }} = {\rm LSTM}_{fw} (\mathbf{w}^{(t)}_{p}, {\mathbf{\overset{\lower0.5em\hbox{$\smash{\scriptscriptstyle\rightarrow}$}}{h^{(t-1)}_p} }}), \\
{\mathbf{\overset{\lower0.5em\hbox{$\smash{\scriptscriptstyle\leftarrow}$}}{h^{(t)}_p} }} = {\rm LSTM}_{bw} (\mathbf{w}^{(t)}_{p}, {\mathbf{\overset{\lower0.5em\hbox{$\smash{\scriptscriptstyle\leftarrow}$}}{h^{(t+1)}_p} }}),
\end{aligned}
\end{equation}
where $\mathbf{w}^{(t)}_{p}$ is the word embedding vector used for the word $w^{(t)}_{p}$.
We introduce a simple encoder architecture that models contexts for synonym discovery, 
which learns to encode the local information around the entity from the raw context without utilizing additional structured annotations.
It focuses on both forward and backward directions. However, the encoding process for each direction ceases immediately after it goes beyond the entity word in the context:
${{\mathbf{h}}_p} = [{\mathbf{\overset{\lower0.5em\hbox{$\smash{\scriptscriptstyle\rightarrow}$}}{\mathbf{h}^{(t_e)}_p} }}, {\mathbf{\overset{\lower0.5em\hbox{$\smash{\scriptscriptstyle\leftarrow}$}}{\mathbf{h}^{(t_e)}_p} }}],$
where $t_e$ is the index of the entity word $e$ in the context and $\mathbf{h}_p\in\mathbb{R}^{1\times d_{CE}}$. By doing this, the context encoder summarizes the context while explicitly considers the entity's location in the context.
Note that more advanced and sophisticated encoding methods can be used, such as ElMo, BERT, or XLNet. The encoder itself is not the main focus of this work.

\subsection{Bilateral Matching with Leaky Unit}
Considering the base case, where we want to identify whether or not two entities, say $e$ and $k$, are synonyms with each other, we propose to find the consensus information from multiple pieces of contexts via a bilateral matching schema.
Recall that for entity $e$, $P$ pieces of contexts $H = \{\mathbf{h}_1, \mathbf{h}_2, ..., \mathbf{h}_{P}\}$ are randomly fetched and encoded. And for entity $k$, we denote $Q$ pieces of contexts being fetched and encoded as $G = \{\mathbf{g}_1, \mathbf{g}_2, ..., \mathbf{g}_Q\}$. 
Instead of focusing on a single piece of context to determine entity synonymity, we adopt a bilateral matching between multiple pieces of encoded contexts for both accuracy and robustness.

\noindent\textit{H$\rightarrow$G} matching phrase:
For each $\mathbf{h_{p}}$ in $H$ and $\mathbf{g_{q}}$ in $G$, the matching score $m_{p \to q}$ is calculated as: 
\begin{equation}
{m_{p \to q}} = \frac{{\exp ({{\mathbf{h}}_p}{{\mathbf{W}}_{{\text{BM}}}}{\mathbf{g}}_q^{\text{T}})}}{{\sum\limits_{p' \in P} {\exp ({{\mathbf{h}}_{p'}}{{\mathbf{W}}_{{\text{BM}}}}{\mathbf{g}}_q^{\text{T}})}}},
\end{equation}
where $\mathbf{W}_{\text{BM}} \in \mathbb{R}^{d_{CE} \times d_{CE}}$ is a bi-linear weight matrix. 

Similarly, the \textit{H$\leftarrow$G} matching phrase considers how much each context $\mathbf{g}_q \in G$ could be useful to $\mathbf{h}_p \in H$:
\begin{equation}
{m_{p \leftarrow q}} = \frac{{\exp ({{\mathbf{g}}_q}{{\mathbf{W}}_{{\text{BM}}}}{\mathbf{h}}_p^{\text{T}})}}{{\sum\limits_{q' \in Q} {\exp ({{\mathbf{g}}_{q'}}{{\mathbf{W}}_{{\text{BM}}}}{\mathbf{h}}_p^{\text{T}})} }}.
\end{equation}
Note that $P\times Q$ matching needs to be conducted in total for each entity pair. We write the equations for each $\mathbf{h}_p\in H$ and ${\mathbf{g}}_q\in G$ for clarity. 
Regarding the implementation, the bilateral matching can be easily written and effectively computed in a matrix form, where a matrix multiplication is used $\mathbf{H}{{\mathbf{W}}_{{\text{BM}}}}\mathbf{G}^T \in \mathbb{R}^{P{\times}Q}$ where $\mathbf{H}\in \mathbb{R}^{P{\times}d_{CE}}$ and $\mathbf{G}\in\mathbb{R}^{Q{\times}d_{CE}}$. The matching score matrix $\mathbf{M}$ can be obtained by taking softmax on the $\mathbf{H}{{\mathbf{W}}_{{\text{BM}}}}\mathbf{G}^T$ matrix over certain axis (over 0-axis for $\mathbf{M}_{p \to q}$, 1-axis for $\mathbf{M}_{p \leftarrow q}$).

Not all contexts are informative during the matching for two given entities. Some contexts may contain
intricate contextual information even if they mention the entity explicitly. In this work, we introduce a leaky unit during the bilateral matching, so that uninformative contexts can be routed via the leaky unit rather than forced to be matched with any informative contexts. 
The leaky unit is a dummy vector $\mathbf{l} \in \mathbb{R}^{1\times d_{CE}}$, where its representation can be either trained with the whole model for each task/dataset, or kept as a fixed zero vector. We adopt the later design for simplicity. 
If we use the \textit{H$\rightarrow$G} matching phrase as an example, the matching score from the leaky unit $\mathbf{l}$ to the $q$-th encoded context in $\mathbf{g}_q$ is:
\begin{equation}
{m_{l \to q}} = \frac{{\exp ({\mathbf{l}}{{\mathbf{W}}_{{\text{BM}}}}{\mathbf{g}}_q^{\text{T}})}}{{\exp ({\mathbf{l}}{{\mathbf{W}}_{{\text{BM}}}}{\mathbf{g}}_q^{\text{T}}) + \sum\limits_{p' \in P} {\exp ({{\mathbf{h}}_{p'}}{{\mathbf{W}}_{{\text{BM}}}}{\mathbf{g}}_q^{\text{T}})} }}.
\end{equation}
If there is any uninformative context in $H$, say the ${\tilde p}$-th encoded context, $\mathbf{h}_{\tilde p}$ will contribute less when matched with $\mathbf{g}_q$ due to the leaky effect: when $\mathbf{h}_{\tilde p}$ is less informative than the leaky unit $\mathbf{l}$. Thus, the matching score between $\mathbf{h}_{\tilde p}$ and $\mathbf{g}_q$ is now calculated as follows:
\begin{equation}
{m_{{\tilde p} \to q}} = \frac{{\exp ({{\mathbf{h}}_{\tilde p}}{{\mathbf{W}}_{{\text{BM}}}}{\mathbf{g}}_q^{\text{T}})}}{{\exp ({\mathbf{l}}{{\mathbf{W}}_{{\text{BM}}}}{\mathbf{g}}_q^{\text{T}}) + \sum\limits_{p' \in P} {\exp ({{\mathbf{h}}_{p'}}{{\mathbf{W}}_{{\text{BM}}}}{\mathbf{g}}_q^{\text{T}})}}}.
\end{equation}

\subsection{Context Aggregation}
The informativeness of a context for an entity should not be a fixed value: it heavily depends on the other entity and the other entity's contexts that we are comparing with.
The bilateral matching scores indicate the matching among multiple pieces of encoded contexts for two entities. For each piece of encoded context, say $\mathbf{g}_q$ for the entity $k$, we use the highest matched score with its counterpart as the relative informativeness score of $\mathbf{g}_q$ to $k$, denote as ${a_q} = \max ({m_{p \to q}}|p \in P)$. Here the intuition is that the informativeness of a piece of context for one entity is characterized by how much it can be matched with the most similar context for the other entity.
We further aggregate multiple pieces of encoded contexts for each entity to a global context based on the relative informativeness scores:
\begin{equation}
\begin{aligned}
 \text{for entity~$e$:}~~~\mathbf{\bar h}  = \sum\nolimits_{p \in P} {{a_p}{{\mathbf{h}}_p}},\\
 \text{for entity~$k$:}~~~\mathbf{\bar g}  = \sum\nolimits_{q \in Q} {{a_q}{{\mathbf{g}}_q}}. 
\end{aligned}
\end{equation}

Note that due to the leaky effect, less informative contexts are not forced to be heavily involved during the aggregation: the leaky unit may be more competitive than contexts that are less informative, thus having a larger matching score. However, as the leaky unit and its matching score are not used for aggregation --scores on informative contexts become more salient during context aggregation.

\subsection{Training Objectives}
We introduce two architectures for training the {\modelname}: a siamese architecture and a triplet architecture.
\\\noindent\textbf{Siamese Architecture}
The Siamese architecture takes two entities $e$ and $k$, along with their contexts $H$ and $G$ as the input. 
The following loss function $L_\text{Siamese}$ is used in training for the Siamese architecture \cite{neculoiu2016learning}:
\begin{equation}
L_\text{Siamese} = y L_{+}(e, k) + (1-y) L_{-}(e, k),
\end{equation}
where it contains losses for two cases: $L_{+}(e,k)$ when $e$ and $k$ are synonyms to each other ($y=1$), and $L_{-}(e,k)$ when $e$ and $k$ are not ($y=0$):
\begin{equation}
\begin{aligned}
  {L_{+}}(e,k) &= \frac{1}{4}{(1 - s(\mathbf{\bar h} ,\mathbf{\bar g} ))^2},\\
  {L_{-}}(e,k) &= {max(s(\mathbf{\bar h} ,\mathbf{\bar g} ) - m,0)^2}, \hfill \\ 
\end{aligned}
\label{eq::siamese}
\end{equation}
where $s(\cdot)$ is a similarity function, e.g. cosine similarity, and $m$ is the margin value. $L_{+}(e,k)$ decreases monotonically as the similarity score becomes higher within the range of [-1,1]. $L_{+}(e,k)=0$ when $s(\mathbf{\bar h} ,\mathbf{\bar g})=1$. For $L_{-}(e,k)$, it remains zero when $s(\mathbf{\bar h} ,\mathbf{\bar g})$ is smaller than a margin $m$. Otherwise $L_{-}(e,k)$ increases as $s(\mathbf{\bar h} ,\mathbf{\bar g})$ becomes larger.

\noindent\textbf{Triplet Architecture}
The Siamese loss makes the model assign rational pairs with absolute high scores and irrational ones with low scores, while the rationality of entity synonymity could be dynamic based on entities and contexts.
The triplet architecture learns a metric such that the global context $\mathbf{\bar h}$ of an entity $e$ is relatively closer to a global context $\mathbf{\bar g_{+}}$ of its synonym entity, say $k_{+}$, than it is to the global context $\mathbf{\bar g_{-}}$ of a negative example $\mathbf{\bar g_{-}}$ by some margin value $m$. The following loss function $L_{{\text{Triplet}}}$ is used in training for the Triplet architecture:
\begin{equation}
{L_{{\text{Triplet}}}} = \max (s({\mathbf{\bar h}},{{{\mathbf{\bar g}}}_ - }) - s({\mathbf{\bar h}},{{{\mathbf{\bar g}}}_ + }) + m,0).
\label{eq::triplet}
\end{equation}

\subsection{Inference}
\begin{figure}[th!]
\centering
\includegraphics[width=0.86\linewidth]{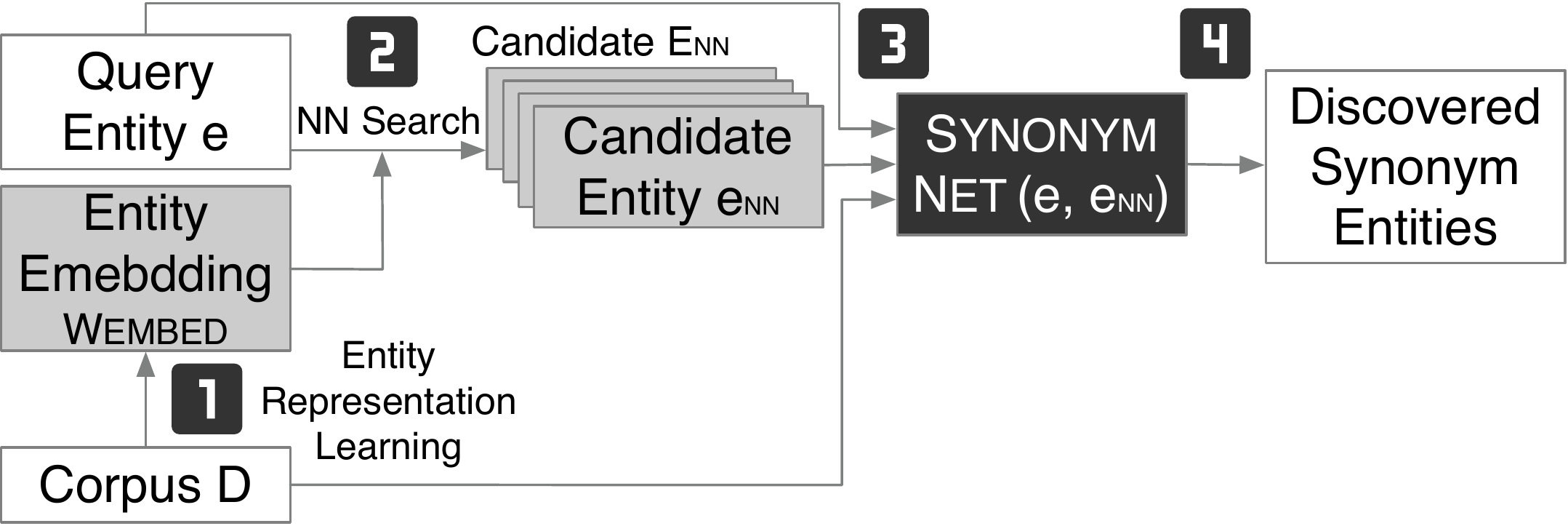}
\caption{Synonym discovery during the inference phase with {\modelname}. 
}\label{fig::framework_overview}
\end{figure}

The objective of the inference phase is to discover synonym entities for a given query entity from the corpus effectively. 
We utilize context-aware word representations to obtain candidate entities that narrow down the search space. The {\modelname} verifies entity synonymity by assigning a synonym score for two entities based on multiple pieces of contexts.
The overall framework is described in Figure \ref{fig::framework_overview}. 

When given a query entity $e$, it is tedious and very ineffective to verify its synonymity with all the other possible entities. 
In the first step, we train entity representation unsupervisely from the massive corpus $D$ using methods such as skipgram \cite{mikolov2013distributed} or GloVe \cite{pennington2014glove}. An embedding matrix can be learned $\mathbf{W}_{\text{EMBED}} \in \mathbb{R} ^{v\times d_{\text{EMBED}}}$, where $v$ is the number of unique tokens in $D$.
Although these unsupervised methods utilize the context information to learn semantically meaningful representations for entities, they are not tailored for the entity synonym discovery task. For example, \texttt{nba championship}, \texttt{chicago black hawks} and \texttt{american league championship series} have similar representations because they tend to share some similar neighboring
words. But they are not synonyms with each other.
However, they do serve as an effective way to obtain candidates because they tend to give entities with similar neighboring context words similar representations. 
In the second step, we construct a candidate entity list $E_{NN}$ by finding nearest neighbors of a query entity $e$ in the entity embedding space of $\mathbb{R}^{d_\text{EMBED}}$. Ranking entities by their embedding proximities with the query entity significantly narrows down the search space for synonym discovery.
In the third step, for each candidate entity $e_{NN} \in E_{NN}$ and the query entity $e$, we randomly fetch multiple pieces of contexts in which entities are mentioned, and feed them into the proposed {\modelname} model.
In the last step, {\modelname} calculates a score $s(e, e_{NN})$ based on the bilateral matching with leaky units over multiple pieces of contexts. The candidate entity $e_{NN}$ is considered as a synonym to the query entity $e$ when it receives a higher score $s(e, e_{NN})$ than other non-synonym entities, or exceeds a specific threshold.

%% file: subfiles/3_experiments.tex
\section{Experiments}
\subsection{Experiment Setup}
\paragraph{Datasets}
Three datasets are prepared to show the effectiveness of the proposed model on synonym discovery. The Wiki dataset contains 6.8M documents from Wikipedia\footnote{https://www.wikipedia.org/} with generic synonym entities obtained from Freebase\footnote{https://developers.google.com/freebase}.
The PubMed is an English dataset where 0.82M research paper abstracts are collected from PubMed\footnote{https://www.ncbi.nlm.nih.gov/pubmed} and UMLS\footnote{https://www.nlm.nih.gov/research/umls/} contains existing entity synonym information in the medical domain. The {Wiki + FreeBase} and {PubMed + UMLS} are public available datasets used in previous synonym discovery tasks \cite{qu2017automatic}.
The MedBook is a Chinese dataset collected by authors where we collect 0.51M pieces of contexts from Chinese medical textbooks as well as online medical question answering forums. Synonym entities in the medical domain are obtained from MKG, a medical knowledge graph.
Table \ref{tab::data_stats} shows the dataset statistics.
\begin{table}[t!]
\centering
\resizebox{0.95\linewidth}{!}{
\begin{tabular}{llll}
\hline
\textbf{Dataset} & \textbf{Wiki + FreeBase} &\textbf{PubMed + UMLS} & \textbf{MedBooK + MKG} \\\hline
\#ENTITY         & 9274         & 6339          & 32,002           \\
$\quad$\#VALID          & 394          & 386     & 661       \\
$\quad$\#TEST           & 104          & 163           & 468       \\
\#SYNSET         & 4615         & 708           & 6600           \\
\#CONTEXT        & 6,839,331    & 815,644       & 514,226       \\
\#VOCAB          & 472,834      & 1,069,061     & 270,027       \\
\#LANGUAGE  & English   & English & Chinese\\
\hline
\end{tabular}
}
\caption{Dataset Statistics.}\label{tab::data_stats}
\end{table}

\paragraph{Preprocessing}
Wiki +Freebase and PubMed + UMLS come with entities and synonym entity annotations, we adopt the Stanford CoreNLP package to do the tokenization. For MedBook, a Chinese word segmentation tool Jieba is used to segment the corpus into meaningful phrases. We remove redundant contexts in the corpus and filter out entities if they appear in the corpus less than five times. For entity representations, the proposed model works with various unsupervised word embedding methods. Here for simplicity, we adopt 200-dimensional word vectors using skip-gram \cite{mikolov2013distributed}. Context window is set as 5 with a negative sampling of 5 words for training.

\paragraph{Evaluation Metrics}
For synonym detection using {\modelname} and other alternatives, we train the models with existing synonym and randomly sampled entity pairs as negative samples. During testing, we also sample random entity pairs as negative samples to evaluate the performance. \textit{Note that all test synonym entities are from unobserved groups of synonym entities: none of the test entities is observed in the training data.} The area under the curve (AUC) and Mean Average Precision (MAP) are used to evaluate the model. A single-tailed t-test is conducted to evaluate the significance of performance improvements when we compare the proposed {\modelname} model with all the other baselines.

For synonym discovery during the inference phase, we obtain candidate entities $E_{NN}$ from K-nearest neighbors of the query entity in the entity embedding space, and rerank them based on the output score $s(e, e_{NN})$ of the {\modelname} for each $e_{NN} \in E_{NN}$. We expect candidate entities in the top positions are more likely to be synonym with the query entity.
We report the precision at position K (P@K), recall at position K (R@K), and F1 score at position K (F1@K). 

\begin{table}[tb!]
\centering
\resizebox{\linewidth}{!}{%
\begin{tabular}{l|cc|cc|cc}\hline
\multirow{2}{*}{\textbf{MODEL}} & \multicolumn{2}{l}{\textbf{Wiki + Freebase}} & \multicolumn{2}{l}{\textbf{PubMed + UMLS}} & \multicolumn{2}{l}{\textbf{MedBook + MKG}} \\\cline{2-3}\cline{4-5}\cline{6-7}
~              & AUC       & MAP      & AUC       & MAP      & AUC       & MAP  \\\hline
word2vec &0.9272 & 0.9371 &0.9301 & 0.9422 & 0.9393 & 0.9418 \\
GloVe  &0.9188 & 0.9295 &0.8890 & 0.8869 & 0.7250 & 0.7049 \\
SRN  &0.8864 & 0.9134 &0.9517 & 0.9559 & 0.9419 & 0.9545 \\
MaLSTM &0.9178 & 0.9413 &0.8151 & 0.8554 & 0.8532   & 0.8833 \\
DPE &0.9461 & 0.9573 &0.9513 & 0.9623 & 0.9479 & 0.9559  \\
\hline
\textbf{\modelname} (Pairwise)          &0.9831$^\dag$ & 0.9818$^\dag$ & \textbf{0.9838}$^\dag$ & \textbf{0.9872}$^\dag$ & \textbf{0.9685} & \textbf{0.9673}\\
\phantom{00} w/o Leaky Unit             &0.9827$^\dag$ & 0.9817$^\dag$ &0.9815$^\dag$ & 0.9847$^\dag$ & 0.9667 & 0.9651  \\
\phantom{00} with Bi-LSTM Encoder   & 0.9683$^\dag$ &0.9625$^\dag$ &0.9495 & 0.9456 & 0.9311& 0.9156 \\\hline
\textbf{\modelname} (Triplet)           &\textbf{0.9877}$^\dag$ & \textbf{0.9892}$^\dag$ &{0.9788}$^\dag$ & 0.9800$^\dag$ & 0.9410 & 0.9230 \\
\phantom{00} w/o Leaky Unit             &0.9705$^\dag$ &0.9631$^\dag$ &0.9779$^\dag$ & {0.9821}$^\dag$ & 0.9359 & 0.9214   \\
\phantom{00} with Bi-LSTM Encoder   &0.9582$^\dag$ &0.9531$^\dag$ & 0.9412 & 0.9288 & 0.9047 & 0.8867  \\\hline     
\end{tabular}%
}
\caption{Test performance in AUC and MAP on three datasets. $\dag$ indicates the significant improvement over all baselines (p $<$ 0.05).}
\label{tab::overall}
\end{table}

\paragraph{Baselines}
We compare the proposed model with the following alternatives.
(1) \textbf{word2vec} \cite{mikolov2013distributed}: a word embedding approach based on entity representations learned from the skip-gram algorithm. We use the learned word embedding to train a classifier for synonym discovery. A scoring function $Score_D(u,v)=x_u\mathbf{W}x_v^T$ is used as the objective.
(2) \textbf{GloVe} \cite{pennington2014glove}: another word embedding approach. The entity representations are learned based on the GloVe algorithm. The classifier is trained with the same scoring function $Score_D$, but with the learned glove embedding for synonym discovery. 
(3) \textbf{SRN} \cite{neculoiu2016learning}: a character-level approach that uses a siamese multi-layer bi-directional recurrent neural networks to encode the entity as a sequence of characters. The hidden states are averaged to get an entity representation. Cosine similarity is used in the objective.
(4) \textbf{MaLSTM} \cite{mueller2016siamese}: another character-level approach. We adopt MaLSTM by feeding the character-level sequence to the model. Unlike SRN that uses Bi-LSTM, MaLSTM uses a single direction LSTM and $l$-1 norm is used to measure the distance between two entities.
(5) \textbf{DPE} \cite{qu2017automatic}: a model that utilizes dependency parsing results as the structured annotation on a single piece of context for synonym discovery.

\subsection{Performance Evaluation}
We report Area Under the Curve (AUC) and Mean Average Precision (MAP) in Table \ref{tab::overall}.
From the upper part of Table \ref{tab::overall} we can see that {\modelname} performances consistently better than those from baselines on three datasets. {\modelname} with the triplet training objective achieves the best performance on Wiki +Freebase, while the Siamese objective works better on PubMed + UMLS and MedBook + MKG. Word2vec is generally performing better than GloVe. SRNs achieve decent performance on PubMed + UMLS and MedBook + MKG. This is probably because the synonym entities obtained from the medical domain tend to share more character-level similarities, such as \texttt{6-aminohexanoic acid} and \texttt{aminocaproic acid}. However, even if the character-level features are not explicitly used in our model, our model still performances better, by exploiting multiple pieces of contexts effectively. DPE has the best performance among other baselines, by annotating each piece of context with dependency parsing results. However, the dependency parsing results could be error-prone for the synonym discovery task, especially when two entities share the similar usage but with different semantics, such as \texttt{NBA finals} and \texttt{NFL playoffs}. 
Table \ref{tab::discovery} reports the performance on Synonym Discovery in P$@$K, R$@$K, and F1$@$K.

We conduct statistical significance tests to validate the performance improvement. The single-tailed t-test is performed for all experiments, which measures whether or not the results from the proposed model are significantly better than ones from baselines. The numbers with $\dag$ markers in Table \ref{tab::overall} indicate that the improvement is significant with
p$<$0.05.

Table \ref{tab::candidate} shows a case for entity \texttt{UNGA}. In the upper part of Table \ref{tab::candidate}, candidate entities are generated with nearest neighbor search on pretrained word embeddings using skip-gram. The lower part of Table \ref{tab::candidate} shows the discovered synonym entities by refining the candidates using the proposed {\modelname} model, where a threshold score of 0.8 is used.
\begin{table}[thb!]
\centering
\centering
\resizebox{0.9\linewidth}{!}{
\begin{tabular}{ll}\hline
\textbf{CANDIDATE ENTITIES} & \textbf{COSINE SIMILARITY} \\\hline
united\_nations\_general\_assembly$||$m.07vp7$||$ &0.847374\\
un\_human\_rights\_council &0.823727\\
the\_united\_nations\_general\_assembly &0.813736\\
un\_security\_council$||$m.07vnr$||$& 0.794973\\
palestine\_national\_council &0.791135\\
world\_health\_assembly$||$m.05\_gl9$||$ &0.790837\\ 
united\_nations\_security\_council$||$m.07vnr$||$ &0.787999\\
general\_assembly\_resolution& 0.784581\\
the\_un\_security\_council &0.784280\\
ctbt &0.777627\\
north\_atlantic\_council$||$m.05pmgy$||$ &0.775703\\
resolution\_1441 &0.773064\\
non-binding\_resolution$||$m.02pj22f$||$& 0.771475\\
unga$||$m.07vp7$||$& 0.770623\\ \hline
\textbf{FINAL ENTITIES} & \textbf{{\modelname} SCORE} \\\hline
united\_nations\_general\_assembly$||$m.07vp7$||$ & 0.842602\\
the\_united\_nations\_general\_assembly&  0.801745\\
unga$||$m.07vp7$||$&  0.800719\\ \hline
\end{tabular}
}
\caption{Candidate entities retrieved using nearest neighbors on Word2vec (upper) and the discovered synonym entities using {\modelname} for \texttt{UNGA} (lower).
}\label{tab::candidate}
\end{table}
\begin{table}[tb!]
\centering
\resizebox{\linewidth}{!}{%
\begin{tabular}{l|lll|lll|lll}
\hline
~ & \multicolumn{3}{l}{\textbf{Wiki + Freebase}} & \multicolumn{3}{l}{\textbf{PubMed + UMLS}} & \multicolumn{3}{l}{\textbf{MedBook + MedKG}} \\ \hline
                 & P@K    & R@K    & F1@K   & P@K    & R@K    & F1@K   & P@K     & R@K     & F1@K    \\\hline
K=1              &  0.3455      & 0.3455       &0.3455        &  0.2400 & 0.0867 & 0.1253 & 0.3051 &	0.2294	& 0.2486 \\
K=5              &  0.1818      & 0.9091       &0.3030        &  0.2880 & 0.7967	& 0.3949  &   0.2388 &	0.8735 &	0.3536    \\
K=10             & 0.1000       & 1.0000       & 0.1818       &   0.1800     &  1.0000  & 0.2915       &  0.1418 & 1.0000 &	0.2360   \\\hline
\end{tabular}
}
\caption{Performance on Synonym Discovery.}\label{tab::discovery}
\end{table}

\paragraph{Ablation Study}
To study the contribution of different modules of {\modelname} for synonym discovery, we also report ablation test results in the lower part of Table \ref{tab::overall}. ``with Bi-LSTM Encoder" uses Bi-LSTM as the context encoder. The last hidden states in both forward and backward directions in Bi-LSTM are concatenated; ``w/o Leaky Unit" does not have the ability to ignore uninformative contexts during the bilateral matching process: all contexts retrieved based on the entity, whether informative or not, are utilized in bilateral matching.
From the lower part of Table \ref{tab::overall} we can see that both modules (Leaky Unit and the Context Encoder) contribute to the effectiveness of the model. The leaky unit contributes 1.72\% improvement in AUC and 2.61\% improvement in MAP on the Wiki dataset when trained with the triplet objective. The Context Encoder gives the model an average of 3.17\% improvement in AUC on all three datasets, and up to 5.17\% improvement in MAP. 

\paragraph{Hyperparameters}
We train the proposed model with a wide range of hyperparameter configurations, as shown in Table \ref{tab::hyperparameters}. 
For the model architecture, we vary the number of randomly sampled contexts $P=Q$ for each entity from 1 to 20. 
Each piece of context is chunked by a maximum length of $T$. For the context encoder, we vary the hidden dimension $d_{CE}$ from 8 to 1024. The margin value $m$ in triplet loss function is varied from 0.1 to 1.75.  
For the training, we try different optimizers, vary batch sizes and 
learning rates. We apply random search to obtain the best-performing hyperparameter setting on the validation dataset, listed in Table \ref{tab::hyper_use}. 
\begin{table}[tbh!]
\centering
\resizebox{0.9\linewidth}{!}{
\begin{tabular}{l|l}\hline
\textbf{HYPERPARAMETERS} & \textbf{VALUE} \\\hline
$P$ (context number) & \{1, 3, 5, 10, 15, 20\} \\
$T$ (maximum context length) & \{10, 30, 50, 80\} \\
$d_{CE}$ (layer size)     & \{8, 16, 32, 64, 128, 256, 512, 1024\}  \\
$m$ (margin)         & \{0.1, 0.25, 0.5, 0.75, 1.25, 1.5, 1.75\}  \\
Optimizer & \{Adam, RMSProp, Adadelta, Adagrad\}\\
Batch Size & \{4, 8, 16, 32, 64, 128\}\\
Learning Rate & \{0.0003, 0.0001, 0.001, 0.01\}\\
\hline
\end{tabular}
}
\caption{Hyperparameter settings.}\label{tab::hyperparameters}
\end{table}
\begin{table}[tbh!]
\centering
\resizebox{.9\linewidth}{!}{%
\begin{tabular}{l|lllllll}\hline
        \textbf{DATASETS} & $P$ & $T$ & $d_{CE}$ & $m$ & Optimizer & Batch & LR \\\hline
        {Wiki + Freebase} & 20 & 50 & 256 & 0.75 & Adam & 16 & 0.0003 \\
        {PubMed + UMLS}   & 20 & 50 & 512 & 0.5 & Adam & 16 & 0.0003 \\
        {MedBook + MKG} & 5 & 80 & 256 & 0.75 & Adam & 16 & 0.0001 \\\hline
\end{tabular}
}
\caption{Hyperparameters.}\label{tab::hyper_use}
\end{table}

\begin{figure}[tbh!]
    \centering
    \minipage{0.3\linewidth}
    \includegraphics[width=\textwidth]{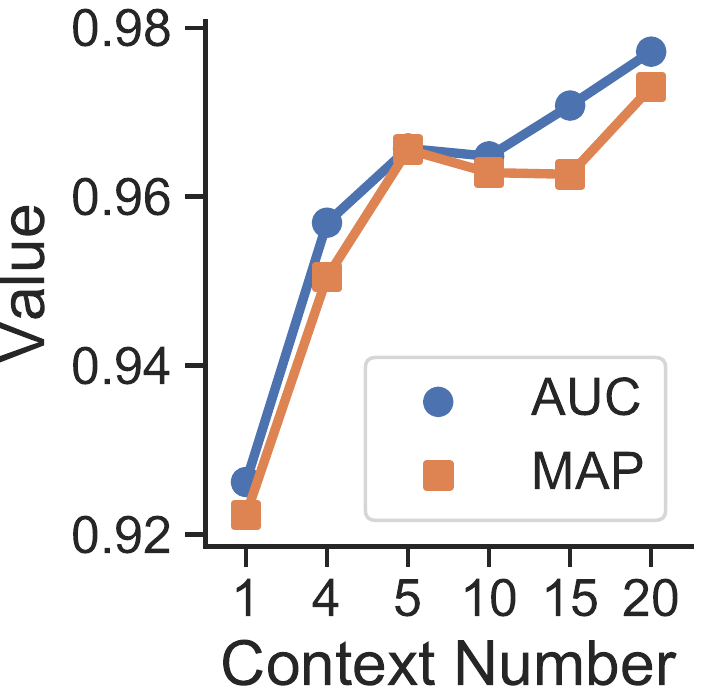}
    \endminipage\hfill
    \minipage{0.3\linewidth} 
    \includegraphics[width=\textwidth]{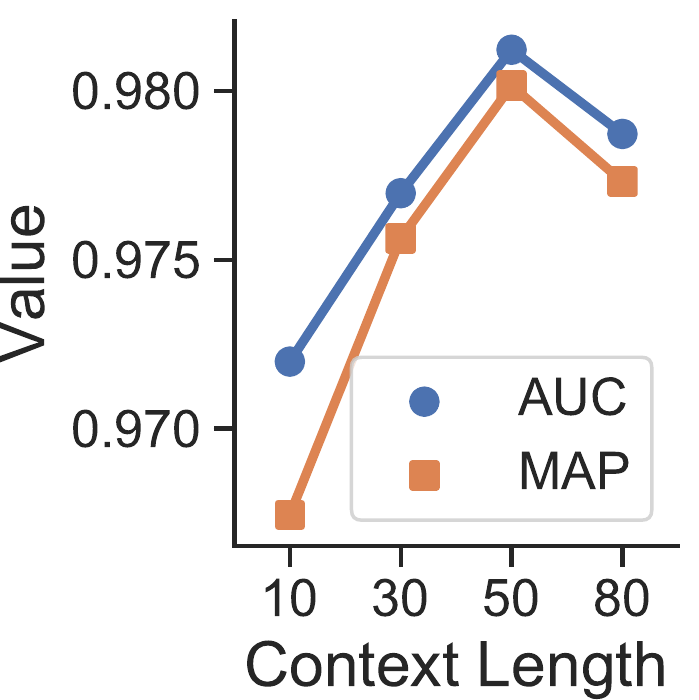}
    \endminipage\hfill
    \minipage{0.4\linewidth}
     \includegraphics[width=\textwidth]{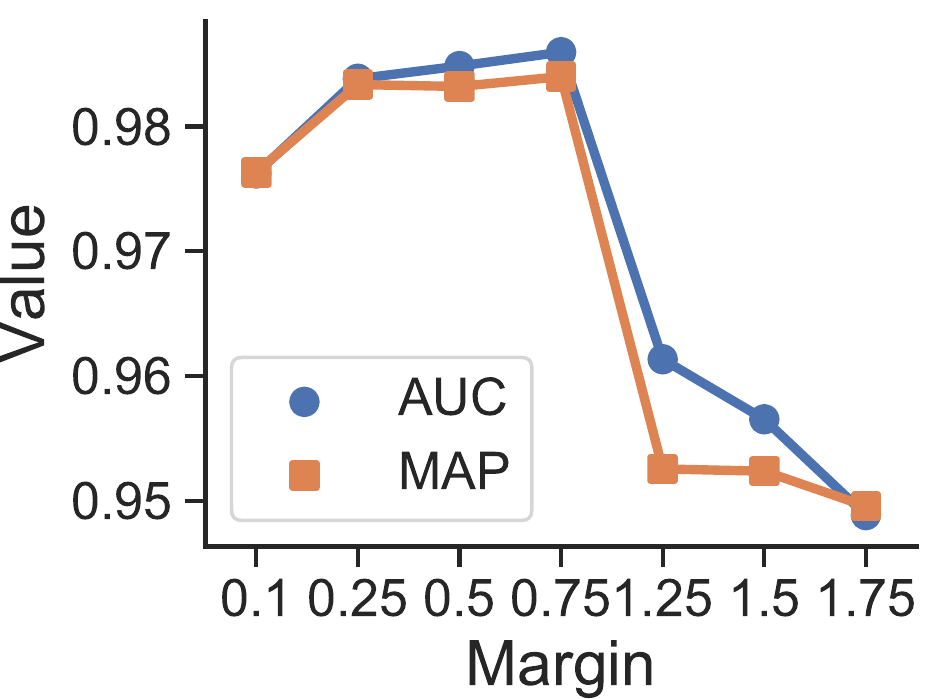}
    \endminipage\hfill
    \caption{Sensitivity analysis.}\label{fig::sensitivity}
\end{figure}

Furthermore, we provide sensitivity analysis of the proposed model with different hyperparameters in Wiki + Freebase dataset in Figure \ref{fig::sensitivity}. Figure \ref{fig::sensitivity} shows the performance curves when we vary one hyperparameter while keeping the remaining fixed. As the number of contexts $P$ increases, the model generally performs better. Due to limitations on computing resources, we are only able to verify the performance of up to 20 pieces of randomly sampled contexts. The model achieves the best AUC and MAP when the maximum context length $T=50$: longer contexts may introduce noise while shorter contexts may be less informative.

%% file: subfiles/4_related.tex
\section{Related Works}
\paragraph{Synonym Discovery} The synonym discovery focuses on detecting entity synonyms.
Most existing works try to achieve this goal by learning from structured information such as query logs \cite{ren2015synonym,chaudhuri2009exploiting,wei2009context}. While in this work, we focus on synonym discovery from free-text natural language contexts, which requires less annotation and is more challenging. 
Some existing works try to detect entity synonyms by entity-level similarities \cite{lin2003identifying,roller2014inclusive,neculoiu2016learning,wieting2016charagram}. For example, distributional features are introduced in \cite{roller2014inclusive} for hypernym detection. Character-level encoding approaches such as \cite{neculoiu2016learning} treat each entity as a sequence of characters, and use a Bi-LSTM to encode the entity information. Such approach may be helpful for synonyms with similar spellings, or abbreviations. Without considering the context information, it is hard for the aforementioned methods to infer synonyms that share similar semantics but are not alike verbatim.
Various approaches \cite{snow2005learning,sun2010semi,liao2017deep,cambria2018senticnet} are proposed to incorporate context information to characterize entity mentions. These models are not designed for synonym discovery.
Dependency parsing result and manually crafted rules on the contexts are used in \cite{qu2017automatic} as the structured annotations for synonym discovery. 
\cite{mudgal2018deep,kasai2019low} assume that entities are given as structured records extracted from texts, where each entity record provides  contextual information about the entity. The goal is to determine whether two entities are the same by comparing and aligning their attributes. We discover synonym entities without such structured annotations.

\paragraph{Sentence Matching}
There is another related research area that studies sentence matching. Early works try to learn a meaningful single vector to represent the sentence \cite{tan2015lstm,mueller2016siamese}.
DSSM style convolution encoders are adopted in \cite{huang2013learning,shen2014learning,palangi2016deep} to learn sentence representations. They utilize user click-through data and learn query/document embeddings for information retrieval and web search ranking tasks.
Although the above methods achieve decent performance on sentence-level matching, the sentence matching task is different from context modeling for synonym discovery in essence. Context matching focuses on local information, while the overall sentence could contain much more information, which is useful to represent the sentence-level semantics, but can be quite noisy for context modeling. 
Matching schemes on multiple instances with varying granularities are introduced in \cite{wang2016compare,wang2016multi,wang2017bilateral}.
However, these models do not consider the word-level interactions from two sentences during the matching. 
Sentence matching models do not explicitly deal with uninformative instances. In context matching, missing such property could be unsatisfactory as noisy contexts exist among multiple contexts for an entity. We adopt a bilateral matching which involves a leaky unit to explicitly deal with uninformative contexts while preserving the expression diversity from multiple pieces of contexts.

%% file: subfiles/5_conclusion.tex
\section{Conclusions}
In this paper, we propose a framework for synonym discovery from free-text corpus in an open-world setting.
A novel neural network model {\modelname} is introduced for synonym detection, which tries to determine whether or not two given entities are synonym with each other. 
{\modelname} makes use of multiple pieces of contexts in which each entity is mentioned, and compares the context-level similarity via a bilateral matching schema to determine synonymity.
Experiments on three real-world datasets show that the proposed method {\modelname} has the ability to discover synonym entities effectively on both generic and domain-specific datasets with an improvement up to 4.16\% in AUC and 3.19\% in MAP. 